\documentclass[]{bytedance_seed}

\usepackage[toc,page,header]{appendix}

\usepackage{minitoc}
\usepackage{enumitem}
\usepackage{pifont}
\usepackage{amssymb}
\usepackage{makecell}
\usepackage{hyperref}
\usepackage{tikzsymbols}
\usepackage{chngcntr}

\usepackage{booktabs}
\usepackage{longtable}

\definecolor{sred}{HTML}{97261c} 
\definecolor{sgreen}{HTML}{266d4a} 

\usepackage{xcolor}
\usepackage{booktabs}

\definecolor{first}{RGB}{200,240,255}    % lightblue

\usepackage{graphicx}
\usepackage{subcaption}
\usepackage{array}

\usepackage{latexsym}
\usepackage{enumitem}
\usepackage{tabularx}
\usepackage{graphicx} 
\usepackage{booktabs}
\usepackage{multirow} 
\usepackage{colortbl} 
\usepackage{tabularx}
\usepackage{makecell}
\usepackage{booktabs}
\usepackage{listings}

\usepackage{wrapfig}
\usepackage{caption}
\usepackage{xspace}
\usetikzlibrary{intersections}
\usepackage{paracol}
\usepackage{lipsum}

\usepackage{tcolorbox}
\tcbuselibrary{breakable, skins}
\usepackage{tikz}
\usepackage{booktabs}
\usepackage{xcolor}
\usepackage{array}

\usepackage{graphicx}
\usepackage{float}
\usepackage{wrapfig}
\usepackage{placeins}

\newtcolorbox[auto counter, number within=section]{Prompt}[2][]{%
  colback=white, % Background color (light sky blue)
  colframe=cyan, % Frame color (sky blue)
  width=\textwidth, % Box width equal to page width
  arc=3mm, 
  boxrule=0.8mm, % Border thickness
  title=\large #2, % Caption text with smaller font size
  breakable, % Supports page breaks
  fonttitle=\small, % Title font size
  fontupper=\footnotesize, % Content font size
  #1 % Additional options
}

\newtcolorbox[auto counter, number within=section]{QuestionCase}[2][]{%
  colback=white,
  colframe=yellow!50!red,
  width=\textwidth, % Box width equal to page width
  arc=3mm, % Sharp corners
  boxrule=0.8mm, % Border thickness
  title=\large #2, % Caption text with smaller font size
  breakable, % Supports page breaks
  fonttitle=\small, % Title font size
  fontupper=\footnotesize, % Content font size
  #1 % Additional options
}

\usepackage{amsmath}
\usepackage{amssymb}  % 提供更多数学符号支持
\usepackage{amsbsy}   % 优化数学字体显示

\usepackage{circledsteps}

\usepackage{mdframed}
\usepackage{listings}
\usepackage{xcolor}

\mdfdefinestyle{promptbox}{
    linecolor=black,
    linewidth=1pt,
    leftmargin=1em,
    rightmargin=1em,
    innertopmargin=1em,
    innerbottommargin=1em,
    innerleftmargin=1em,
    innerrightmargin=1em,
    backgroundcolor=gray!5,
    roundcorner=5pt
}

\abstract{
As Large Language Models (LLMs) exhibit plateauing performance on conventional benchmarks, a pivotal challenge persists: evaluating their proficiency in complex, open-ended tasks characterizing genuine expert-level cognition. Existing frameworks suffer from narrow domain coverage, reliance on generalist tasks, or self-evaluation biases. To bridge this gap, we present XpertBench, a high-fidelity benchmark engineered to assess LLMs across authentic professional domains.

XpertBench consists of 1,346 meticulously curated tasks across 80 categories, spanning finance, healthcare, legal services, education, and dual-track research (STEM and Humanities). These tasks are derived from over 1,000 submissions by domain experts--including researchers from elite institutions and practitioners with extensive clinical or industrial experience--ensuring superior ecological validity. Each task uses detailed rubrics with mostly 15-40 weighted checkpoints to assess professional rigor. To facilitate scalable yet human-aligned assessment, we introduce ShotJudge, a novel evaluation paradigm that employs LLM judges calibrated with expert few-shot exemplars to mitigate self-rewarding biases.

Our empirical evaluation of state-of-the-art LLMs reveals a pronounced performance ceiling: even leading models achieve a peak success rate of only ~66\%, with a mean score around 55\%. Models also exhibit domain-specific divergence, showing non-overlapping strengths in quantitative reasoning versus linguistic synthesis.. These findings underscore a significant "expert-gap" in current AI systems and establish XpertBench as a critical instrument for navigating the transition from general-purpose assistants to specialized professional collaborators.                                                                                                                                                                                          

}

\begin{document}
\title{Xpertbench: Expert Level Tasks with Rubrics-Based Evaluation}

% \affiliation[1]{ByteDance Seed}
% \affiliation[2]{xxx}
% \contribution[*]{Equal contribution}
% \contribution[\dagger]{Corresponding authors}
% \contribution[\ddagger]{Work done at ByteDance Seed}

\affiliation[]{ByteDance Seed}

\contribution{Full author list in Contributions}

\maketitle

% You can add additional info fields as follows 

\checkdata[Project Page]{}
\section{Introduction}

As Large Language Models (LLMs) evolve from passive QA systems into autonomous agents, current evaluation paradigms are increasingly exposing their limitations. Traditional "exam-style" benchmarks (e.g., MMLU-Pro \cite{mmlupro}, GPQA \cite{gpqa}) provide easy verifiability but suffer from rapid saturation. Recent efforts to mitigate this have largely focused on raising the difficulty ceiling, curating extreme edge-cases \cite{humanitylastexam} or unsolved mathematical problems \cite{frontiermath}. Yet, scaling difficulty within a closed-form paradigm still reduces evaluation to isolated questions with singular answers. Even benchmarks targeting agentic capabilities and deep web retrieval, such as GAIA \cite{gaia} and BROWSECOMP \cite{browsecomp}, ultimately collapse complex, multi-step research into short factoids or specific reference strings. By flattening open-ended synthesis and professional judgment into point-estimate metrics, these frameworks maintain a severe disconnect between empirical scores and practical utility. Therefore, it is imperative that the field transcends static knowledge testing and reorients towards evaluating end-to-end, authentic tasks that mirror expert-level workflows as LLMs are increasingly integrated as professional co-pilots.

To bridge the disconnect between empirical scores on traditional benchmarks and the practical utility of AI systems, we introduce XpertBench, a high-fidelity benchmark explicitly engineered to evaluate LLMs on end-to-end, real-world expert workflows. To ensure superior ecological validity, XpertBench is constructed driven by three core characteristics:
\begin{itemize}
\item \textbf{Open-Ended, Long-Horizon Tasks:} Diverging from closed-form, "exam-style" paradigms that primarily test static knowledge recall, XpertBench focuses on tasks akin to deep research. Genuine expert problem-solving is inherently ill-structured; it requires navigating ambiguity, synthesizing extensive domain-specific literature, and resolving conflicting constraints—capabilities that point-estimate metrics completely fail to capture.
\item \textbf{High-Stakes, Comprehensive Domain Coverage:} We anchor our evaluation in seven professional domains (e.g., Finance, Law, Healthcare, Education) chosen for their substantial economic contribution, high cognitive complexity, and significant societal impact. Compared to recent efforts like \$OneMillion-Bench and GDPval, XpertBench not only significantly scales up the volume of tasks but uniquely incorporates historically underrepresented yet critical fields such as Education (24.4\%) and Humanities \& Social Sciences (8.6\%), making it significantly more persuasive in evaluating "generalist" professional capabilities.
\item \textbf{Elite Expert Curation and Granular Rubrics:} We implemented a highly rigorous, expert-centric curation pipeline engaging over 1,000 elite domain experts (e.g., active researchers, CFAs, CPAs, MDs, JDs). Following a stringent two-stage qualification, these experts meticulously reconstructed their daily professional challenges into 1,346 testable scenarios. After multi-stage peer-review filtering to eliminate subjective edge cases, every single task is underpinned by an objective, multi-faceted evaluation rubric featuring 15–40 granular checkpoints.
\end{itemize}

Our empirical evaluation of 12 state-of-the-art models across this benchmark yields profound insights into the true boundaries of frontier AI. Overall, the Claude and GPT model families distinctly separate themselves from the pack, delivering the most robust and practically viable expert-level experiences. However, a closer inspection reveals significant capability gaps: model performance noticeably deteriorates on end-to-end workflows in the STEM and Education domains, where rigid formal logic and long-horizon pedagogical planning are paramount. Beyond top-line scores, our fine-grained behavioral analysis exposes critical failure modes in how current models falter. Rather than making simple factual errors, models frequently suffer from retrieval interference---where persistent web browsing introduces extraneous noise that distracts from the core analytical trajectory, severely degrading ultimate usability. Furthermore, models exhibit severe principle hallucinations; a fundamental conceptual misstep early in a task often cascades, rendering the entire subsequent reasoning chain logically incoherent and practically unusable. Finally, our analysis highlights pronounced domain-specific specialization: for instance, GPT-5.4-high overwhelmingly dominates Finance (84.65\%) but lags in STEM (42.84\%), whereas Claude-Opus-4.6-thinking excels in Law and Humanities. This scale of evidence conclusively demonstrates that a singular ``omni-capable'' expert model does not yet exist.

In summary, our core contributions are three-fold:
\begin{itemize}
    \item \textbf{A High-Fidelity Benchmark:} We release XpertBench, an unprecedented, multi-domain benchmark that significantly expands the scale, coverage, and depth of expert-level evaluation, serving as a critical instrument for measuring real-world AI utility.
    \item \textbf{A Robust Evaluation Pipeline:} We formalize a fixed, high-quality methodological pipeline---from expert-driven task curation and dual-weighted atomic rubrics to the ShotJudge paradigm---establishing a scalable, human-aligned standard for end-to-end generative assessment.
    \item \textbf{Critical Diagnostics of Frontier Models:} We provide deep empirical insights into the behavioral flaws of leading LLMs, demonstrating how non-overlapping domain expertise, retrieval interference, and reasoning hallucinations currently bottleneck the transition from general assistants to specialized professional co-pilots.
\end{itemize}

\begin{figure}[h]
    \centering
    \includegraphics[width=\textwidth]{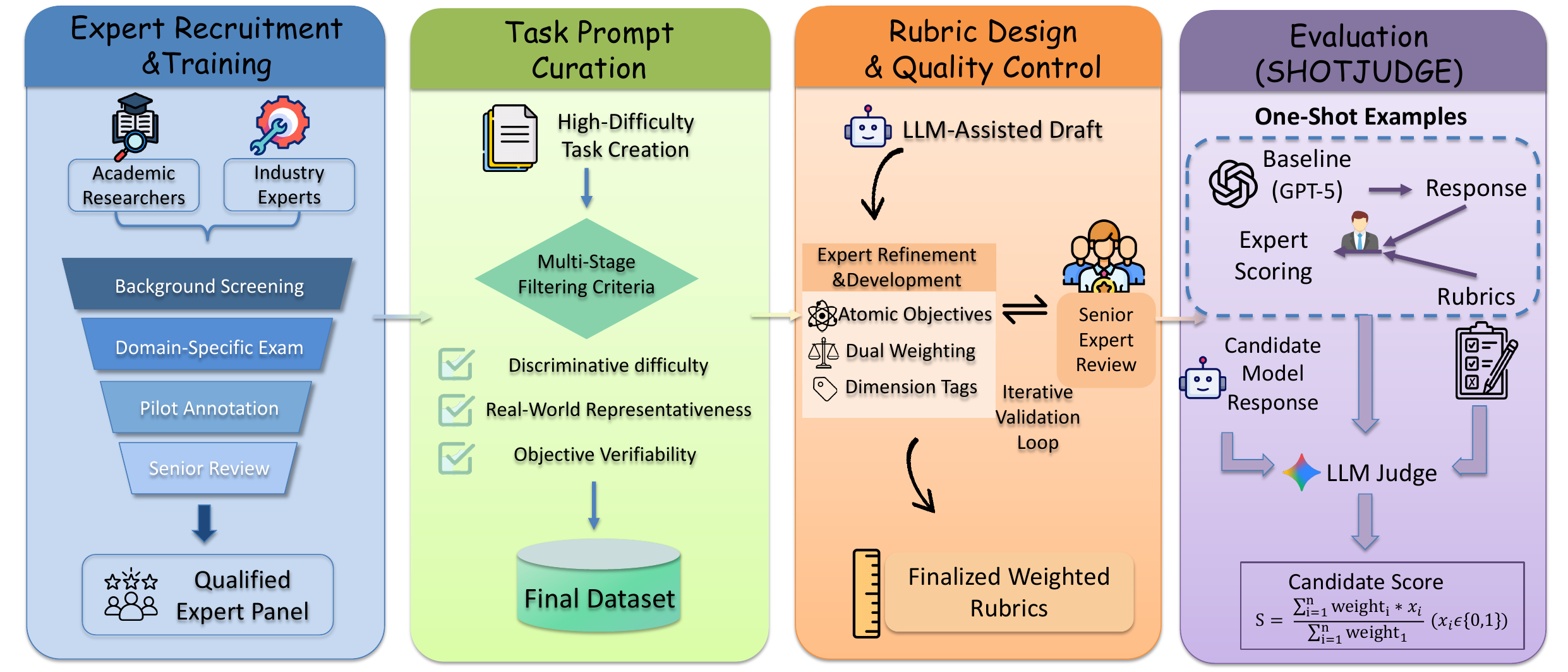}
    \vspace{-0.2in}
    \caption{
    Overview of the benchmark construction and evaluation pipeline, consisting of expert recruitment, task curation, rubric design, and the SHOTJUDGE evaluation framework.
    }
    \vspace{-0.2in}
    \label{fig:pipeline}
\end{figure}

\section{Related Work}
\subsection{Expert-level-task}
Consequently, a variety of specialized benchmarks have emerged to assess expertise in narrow professional fields. For instance, MedQA and PubMedQA \cite{medqa,pubmedqa} evaluate medical knowledge, SciBench \cite{scibench} and LegalBench \cite{legalbench} assess technical problem-solving in STEM and law, and FinBen \cite{finben} provides a comprehensive suite of financial tasks. However, while these benchmarks are valuable within their specific disciplines, their isolated nature fails to measure the cross-domain synthesis and adaptable reasoning essential for real-world, versatile AI assistants.

Recent benchmarks have sought to address earlier limitations by increasing task difficulty and domain coverage. For instance, MMLU-Pro \cite{mmlupro} and GPQA \cite{gpqa} evaluate deep domain knowledge, while Humanity's Last Exam (HLE) \cite{humanitylastexam} introduces 2,500 expert-level questions where frontier models still score below 10\% accuracy. FrontierMath \cite{frontiermath} targets research-level mathematical reasoning with problems so challenging that leading models solve less than 2\% of them. Concurrently, AgentBench \cite{agentbench} and GAIA \cite{gaia} focus on multi-step task execution. Despite these advances, a fundamental limitation persists: these frameworks still rely on an "exam-based" paradigm designed around well-defined questions with verifiable answers. Such tasks primarily test knowledge recall or advanced retrieval. Consequently, they fail to capture the ill-structured, open-ended nature of real-world expert problem-solving, which frequently requires navigating ambiguity and resolving conflicting constraints.

Concurrently, the emergence of Deep Research Agents has prompted a new line of benchmarks focused on complex, open-ended research tasks. BrowseComp \cite{browsecomp} evaluates agents' ability to locate hard-to-find information through persistent web browsing, while DeepResearch Bench \cite{deepresearchbench} provides 100 PhD-level research tasks across 22 fields with dual evaluation frameworks (RACE and FACT). DEER \cite{deer} proposes a comprehensive evaluation framework for deep research reports using expert-systematized rubric items and exhaustive claim-level fact-checking. These benchmarks represent an important shift toward evaluating generative, research-oriented capabilities, yet they primarily target the information retrieval and synthesis pipeline rather than the domain-specific professional judgment that characterizes expert practice.

XpertBench diverges from these precedents by prioritizing ecological validity. Rather than relying on academic proxies, we source 1346 tasks directly from the workflows of active practitioners across seven high-demand domains. By anchoring our benchmark in authentic professional challenges, we transcend the knowledge-retrieval bottleneck and provide a more rigorous assessment of LLMs’ ability to function within the complex heuristics of real-world professional practice.

\subsection{Rubrics evaluated benchmark}
As LLM outputs shift from short-form answers to complex, generative artifacts, traditional metrics like Exact Match or ROUGE have become increasingly obsolete. This has necessitated a shift toward rubric-based evaluation, which decomposes holistic quality into granular, interpretable dimensions.

Existing literature primarily bifurcates into two evaluative extremes. On one end, automated "LLM-as-a-judge" frameworks, such as AlpacaEval \cite{alpacaeval}, MT-Bench \cite{mtbench}, Arena-Hard \cite{arenahard} and WildBench \cite{wildbench} offer unparalleled scalability but are frequently criticized for methodological circularity. WildBench further advances this line by sourcing challenging tasks from real users and employing task-specific checklists, yet still relies on LLM judges for scoring. Furthermore, while recent frameworks like Prometheus \cite{bi2024prometheus} introduce fine-grained rubric learning, they still face challenges in generalizing beyond their synthetic training distributions. JudgeBench \cite{judgebench} explicitly evaluates the reliability of LLM-based judges themselves, revealing that even strong models like GPT-4o perform only marginally better than random guessing on challenging response pairs---underscoring the fragility of automated evaluation. When models are evaluated against AI-generated criteria, there is an inherent risk of "self-enhancement bias," where the judge rewards stylistic alignment with its own training distribution rather than authentic professional merit. On the other end, human-centric frameworks like HELM \cite{helm} or ChatbotArena \cite{chatbotarena} offer higher fidelity but are constrained by logistical and financial bottlenecks, particularly when the tasks require highly specialized (e.g., PhD or JD-level) expertise.

Furthermore, existing rubrics often suffer from a granularity deficit. Many frameworks employ coarse-grained Likert scales or generic criteria (e.g., "helpfulness," "coherence") that fail to capture the nuanced technical requirements of professional-grade outputs. While recent works like SimpleQA \cite{simpleqa} attempt to re-introduce factuality-based rigor, they often sacrifice the multi-dimensional assessment necessary for open-ended expert tasks. RubricEval \cite{rubriceval} proposes a human-LLM hybrid framework where experts generate instruction-level criteria and LLM evaluators score against detailed rubrics, demonstrating improved alignment but remaining limited to general-purpose instructions rather than domain-specific professional standards.

To reconcile the tension between evaluative rigor and scalability, we introduce ShotJudge. Unlike prior methods that rely on "black-box" AI judgments, our methodology is underpinned by expert-anchored rubrics comprising 15-40 granular, weighted checkpoints per task. By utilizing human-expert assessments as few-shot calibration anchors, ShotJudge ensures that the automated judge remains grounded in professional standards rather than latent model biases. This hybrid approach allows XpertBench to maintain the precision of expert peer review while achieving the throughput required for large-scale model benchmarking.

\section{Overview of XpertBench}

We introduce \textit{XpertBench}, a multi-domain benchmark designed to evaluate the true performance boundaries of Large Language Models (LLMs) in high-value, open-ended, and long-horizon tasks. The dataset comprises 1,346 complex tasks that require models to demonstrate capabilities ranging from strategic planning and logical deduction to professional judgment and cultural interpretation.

The composition of XpertBench is driven by a rigorous domain selection process targeting knowledge-intensive sectors characterized by three pivotal criteria: substantial economic contribution, high cognitive complexity, and significant societal impact. As detailed in Table~\ref{tab:benchmark_stats}, the dataset spans seven distinct professional domains, selected to mirror the operational realities of the modern tertiary sector. So sectors characterized by high cognitive complexity and significant societal impact is targeted. We prioritize \textbf{Finance} and \textbf{Law} to evaluate capabilities in high-stakes decision-making, requiring precise statutory interpretation and risk assessment essential for economic stability. Complementing these are the \textbf{Education} and \textbf{Medical} sectors, which address critical dimensions of public welfare by testing pedagogical planning and clinical judgment in developmental and life-critical contexts. Finally, to capture the full spectrum of theoretical reasoning, the benchmark incorporates \textbf{STEM}, \textbf{Computer Science \& Mathematics}, and \textbf{Humanities \& Social Sciences (HSS)}; this combination assesses diverse intellectual demands, ranging from the rigorous logical deduction required in applied sciences to the nuanced cultural interpretation central to the humanities.

\begin{figure}[H]
    \centering
    \includegraphics[width=\textwidth]{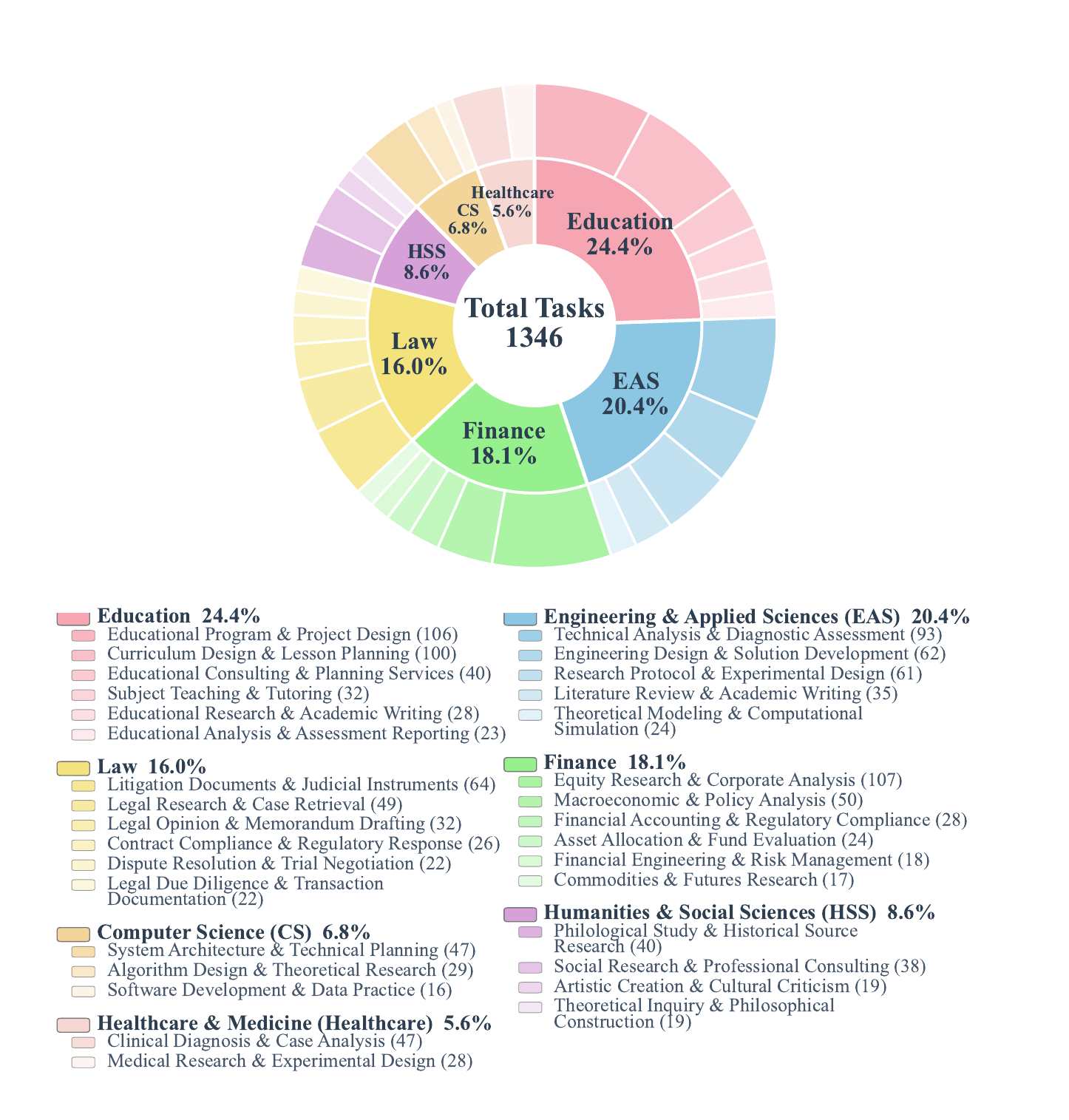}
    \vspace{-0.2in}
    \caption{
    Distribution of tasks across professional domains in XpertBench, highlighting the balance between STEM and Humanities fields.
    }
    \vspace{-0.2in}
    \label{fig:domain_dist}
\end{figure}

\begin{table}[t]
\centering
\caption{Detailed Statistics of XpertBench across Professional Domains. EAS stands for Engineering and Applied Sciences; HSS stands for Humanities and Social Sciences. Healthcare refers to Healthcare \& Medicine.}
\label{tab:benchmark_stats}
\renewcommand{\arraystretch}{1.2}
\small
\begin{tabular}{lcccc}
\toprule
\textbf{Domain} & \textbf{\# Tasks} & \textbf{\# of Rubrics} & \textbf{Prompt Length} & \textbf{Primary Competency} \\ 
\midrule
Finance & 244 & 17 & 533 & Market Analysis \\
Law & 215 & 18 & 413 & Juridical Reasoning \\
Education & 329 & 15 & 960 & Pedagogy \& Planning \\
EAS & 275 & 19 & 699 & Technical Mechanism \\
HSS & 116 & 15 & 373 & Cultural Interpretation \\
Computer Science & 92 & 17 & 558 & Logical Deduction \\
Healthcare & 75 & 16 & 371 & Clinical Judgment \\
\midrule
\textbf{Total / Average} & \textbf{1346} & \textbf{17} & \textbf{631} & \textbf{Cross-Domain Expertise} \\
\bottomrule
\end{tabular}
\end{table}

\subsection{Expert Recruitment and Training}
Experts are sourced through the Xpert Data Platform. To ensure annotation quality align with the complex domains of XpertBench, the expert pool—sourced through a hybrid approach of direct registration, peer referrals, and vendor partnerships—primarily consists of two profiles: active researchers from top-tier universities and seasoned practitioners with substantial industry experience. Consequently, the expert panel is characterized by distinguished academic and professional credentials. Academically, approximately 61\% of the contributors are affiliated with China's premier research institutions (Project 985 and 211) and elite domain-specific universities, supplemented by more than two hundred experts from globally renowned overseas institutions. Professionally, every member possesses a minimum of three years of practical experience. The team's domain authority is further reinforced by high-value industry certifications, including 183 legal professional qualifications, 163 medical practitioner licenses, and a substantial presence of CFA, CPA, and CATTI holders.

Prior to formal participation, each expert undergo a two-stage rigorous qualification process: (1) completing a domain-specific proficiency examination, and (2) candidates performed a trial annotation (or task authoring) under the same instructions that govern the full data-collection pipeline. Trial outputs are reviewed by at least one senior domain annotator; submissions that fail to meet the review standards are returned for revision or rejected. This process ensures that all experts thoroughly understand task requirements and maintain consistent annotation standards. Experts demonstrating insufficient proficiency or inconsistent quality are excluded from the final annotation team.

\subsection{Prompt Curation}
Following the qualification process, we engage trained experts to contribute authentic task prompts derived from their real-world professional scenarios. To ensure alignment with the benchmark's rigorous standards, all experts undergo a standardized training protocol prior to task creation. This training explicitly distinguish between academic ``exam questions'' and authentic ``professional tasks,'' guiding experts to formulate complex, open-ended scenarios that mirror actual workflows rather than rote knowledge retrieval. Experts are instructed to design tasks that challenge the reasoning frontiers of state-of-the-art models, focusing on scenarios with low estimated pass rates to ensure the benchmark maintains high discriminative power and avoids saturation.

To ensure dataset diversity, each expert is limited to submitting a maximum of three prompts. Experts are instructed to provide comprehensive descriptions encompassing three key components: (1) the specific work scenario or context, (2) relevant background materials or domain knowledge required, and (3) detailed output requirements or deliverables. In addition, each expert is required to supply a reference answer illustrating the expected standard of professional reasoning and output quality. This structured approach ensures that collected prompts closely reflect genuine professional demands and maintain ecological validity. 

From an initial pool of over 1,000 submitted prompts, we implement a rigorous multi-stage selection process conducted by our internal panel of domain experts. We filter prompts that pose substantial challenges to current state-of-the-art LLMs, ensuring the benchmark effectively distinguishes model capabilities and avoids ceiling effects. Prompts are also required to represent typical and high-frequency tasks within each professional domain, rather than edge cases or overly specialized scenarios. By examining both the prompt specifications and expert-provided reference answers, we retain only those prompts with clearly definable success criteria and objective evaluation protocols, excluding items that relied on subjective preference. Through this systematic quality control process, we curate a final dataset of 1346 prompts spanning over 80 distinct task categories, achieving both breadth across domains and depth within specialized areas.

\subsection{Rubric Design and Quality Control}

For each curated task, we adopt a structured, rubric-anchored evaluation framework to ensure that model performance can be assessed with high granularity, objectivity, and reproducibility. 

\textbf{LLM-Assisted Rubric Generation.}  We first employ state-of-the-art LLMs (Claude Opus 4.1 or Gemini 2.5 Pro) to generate initial evaluation rubrics based on the prompt specifications and expert-provided reference answers. This LLM-assisted approach provides a structured foundation for experts to refine and develop comprehensive assessment criteria, ensuring consistency while reducing manual effort.

\textbf{Expert-Driven Criteria Development.} Building upon the LLM-generated drafts, the original prompt contributors or additional annotation experts are tasked with crafting detailed evaluation criteria (hereafter referred to as "rubrics") . The majority of tasks comprise 15-40 checkpoints, with each checkpoint designed to capture a specific, essential element of high-quality model responses. To ensure atomicity and objectivity, we enforce the following design principles:

\begin{itemize}
    \item \textit{Granularity}: Each checkpoint must focus on a single, well-defined requirement and cannot conflate multiple evaluation aspects.
    \item \textit{Objectivity}: Checkpoints must be formulated as objective statements that can be unambiguously labeled as TRUE (satisfied) or FALSE (unsatisfied) based on model outputs.
    \item \textit{Specificity}: Checkpoints must explicitly specify the concrete content, format, or reasoning elements that model responses should contain.
\end{itemize}

\textbf{Checklist Weighting and Tagging.} To reflect the varying importance of different evaluation criteria, experts assign dual-level weights to each checkpoint. At the qualitative level, checkpoints are categorized as \textit{Essential}, \textit{Important}, or \textit{Optional}, indicating their relative impact on overall assessment quality. Complementing this categorical classification, experts provide quantitative weights ranging from 1 to 10 based on professional judgment of each checkpoint's relative importance, without predefined distribution constraints. This dual-weighting scheme enables both coarse-grained prioritization and fine-grained score calibration during evaluation.

Additionally, experts classify each checkpoint into predefined evaluation dimensions. For example,  (e.g., factual accuracy, logical coherence, domain-specific expertise) to facilitate multi-faceted analysis of model capabilities(See table \ref{tab:rubric_dimensions}).

\begin{table}[t]
\centering
\caption{Comprehensive Evaluation Rubric Dimensions. This table outlines the specific dimensions and their corresponding detailed descriptions used to assess and score candidate model responses.}
\label{tab:rubric_dimensions}
\renewcommand{\arraystretch}{1.2}
\resizebox{\textwidth}{!}{ % Ensure the table fits within the text width
\begin{tabular}{lp{0.75\linewidth}}
\toprule
\textbf{Dimension} & \textbf{Description} \\
\midrule
Instruction Following & Accurately identifying and understanding task instructions, requirements, and goals. \\
Constraint Satisfaction & Strictly adhering to explicit constraints and requirements stipulated in the prompt. \\
Accuracy \& Factual Correctness & Precision and integrity of domain-specific concepts, theories, regulations, and factual data. \\
Information Processing & Ability to extract, synthesize, and integrate information from literature or given contexts. \\
Logical Coherence & Rigor of reasoning, completeness of argumentation, and sound analysis of causal relationships. \\
Domain Expertise & Application of core industry mindsets, critical thinking, analytical methods, tools, and theoretical frameworks. \\
Completeness \& Executability & Wholeness and practical feasibility of the proposed business or research processes. \\
Normativity \& Compliance & Adherence to regulatory requirements, academic integrity standards, and industry norms. \\
Safety \& Risk Management & Ability to identify risks, control hazards, and analyze potential errors or biases. \\
Novelty \& Creativity & Generation of novel insights, innovative methods, and unique perspectives. \\
Linguistic Quality & Accuracy, professionalism, fluency, and appropriateness of language expression. \\
Structure \& Formatting & Organization of content structure and effectiveness of visual presentation. \\
Depth of Analysis & Excavation of fundamental problem essences and explanation of underlying mechanisms. \\
Interdisciplinary Integration & Synthesis across diverse domains and comprehensive application of multiple perspectives. \\
Information Richness & Inclusion of value-added, supplementary, or beneficial information beyond the core requirement. \\
Other & Miscellaneous relevant factors not covered by other dimensions. \\
\bottomrule
\end{tabular}
}
\end{table}

\textbf{Quality Control.} To ensure rubric quality, each completed set of criteria undergo review by at least one additional expert in the same domain. Reviewers examine the factual validity, clarity, atomicity, and weight assignments of every criterion, providing line-level revision requests where needed. Criteria failing to meet rubric standards would be returned to the annotators for revision; this process continue iteratively until all issues are resolved. In addition, approximately 30\% of tasks underwent internal spot-checks by senior domain experts, who further evaluate rubric consistency, task alignment, and evaluability. Tasks with irreparable issues are either substantially revised or discarded.

\section{Evaluation}

To facilitate a scalable yet expert-aligned assessment, we utilize \textbf{ShotJudge}, an evaluation paradigm that grounds automated scoring in human-expert reasoning via few-shot in-context learning. Recognizing the limitations of zero-shot LLM-as-a-judge approaches—specifically their tendency toward stylistic bias\cite{wang2023large}—ShotJudge employs expert-annotated exemplars to calibrate the judge's evaluative threshold. In our implementation, we employ GPT-5 as the anchor baseline model and Gemini 2.5 Pro as the primary LLM Judge. The methodology is operationalized through a two-stage process: expert anchoring and calibrated automated scoring.

\textbf{Expert Anchoring and Meta-Evaluation.} The foundational stage involves generating "gold-standard" evaluative rationales. A primary cohort of domain experts conducts a rigorous blind review of responses generated by the baseline model. Following the rubric dimensions established in $\S 3.1$, experts provide binary judgments ($s \in \{0, 1\}$) for each criterion, accompanied by a detailed qualitative justification (rationale). This step ensures that the evaluative signal captures not only the correctness of the output but also the professional nuance required by the task. 

To ensure the reliability of these anchors, a secondary cohort of senior experts conducts a cross-verifying meta-evaluation of the initial annotations. This hierarchical review process filters out idiosyncratic biases, producing a robust set of expert-annotated ground truths that serve as the "logical blueprint" for the automated judge.

\textbf{One-Shot Calibrated Scoring.} In the automated phase, the LLM judge is provided with a prompt context containing: (i) the original task prompt, (ii) the expert-designed rubric, and (iii) the baseline model's response paired with its corresponding expert-validated rationales and scores as one-shot exemplars. 

By leveraging these exemplars, the LLM judge is instructed to emulate the expert's reasoning patterns when evaluating a candidate model's response. For each criterion $c_i$, the judge outputs a binary score $x_i \in \{0, 1\}$. The final performance metric is derived via a weighted aggregation:
\begin{equation}
    S = \frac{\sum_{i=1}^{n} w_i x_i}{\sum_{i=1}^{n} w_i}
\end{equation}
where $w_i$ represents the expert-assigned weight for the $i$-th criterion. This mechanism ensures that the final score is an interpretable reflection of professional standards, effectively bridging the gap between human expertise and automated scalability.

\textbf{Human-AI Alignment.} To validate the reliability of ShotJudge, we conduct an alignment analysis between the LLM judge and human experts. We adopt the Consistency minus Discordance Rate (CDR) as the primary reliability metric, defined as $P(\text{agree}) - P(\text{disagree})$. This metric specifically penalizes "rank reversals" where the judge's preference contradicts the expert's. Our results show that ShotJudge achieves a CDR of 52.0\%, significantly outperforming standard zero-shot LLM-as-a-judge baselines. This robust alignment demonstrates that the one-shot calibration effectively transmits expert evaluative intent to the automated system. Due to the extensive human labor required for expert anchoring and meta-evaluation, we construct a highly curated XpertBench-Gold subset (N=245) via stratified sampling across domains. This subset serves as the primary testbed for our empirical evaluation using ShotJudge.

\section{Experiments and Results}

\begin{figure}[H]
    \centering
    \includegraphics[width=\textwidth]{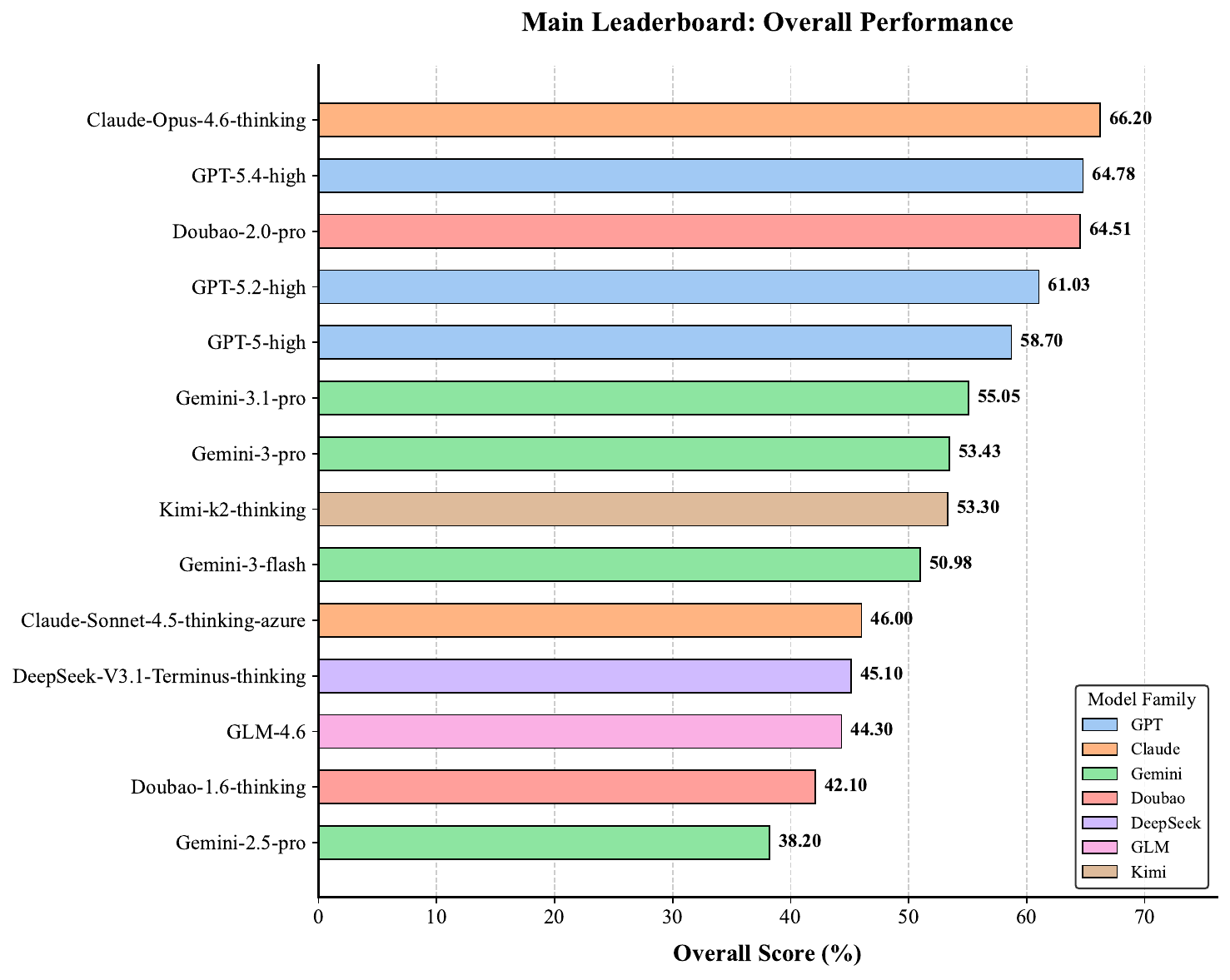}
    %{fig1_main_leaderboard_0325_without_gemini.pdf}
    \vspace{-0.2in}
    \caption{
    Results on XpertBench-Gold evaluation subset (N=245).
    }
    %\vspace{-0.2in}
    \label{fig:overall_performance}
\end{figure}

We conduct a comprehensive evaluation of several state-of-the-art large language models—namely Claude-Opus-4.6-thinking, GPT-5.4-high, Doubao-2.0-pro, and Gemini-3.1-pro across the XpertBench-Gold subset (N=245). Because the rigorous expert calibration process limits the sample sizes for Computer Science and Healthcare in this current gold subset, we focus our fine-grained domain analysis on the five primary domains, while retaining all 245 tasks for the overall performance metric. As depicted in Figure~\ref{fig:overall_performance}, \textbf{Claude-Opus-4.6-thinking} impressively achieves the dominant position with an overall score of 66.20\%. This remarkable performance establishes a new state-of-the-art baseline, with GPT-5.4-high (64.78\%) and Doubao-2.0-pro (64.51\%) following closely behind.

A granular analysis of domain-specific performance (Table~\ref{tab:domain_breakdown}) reveals significant distinctions in model capabilities similar to the trade-offs observed between accuracy and aesthetics in prior works. While top-tier models excel significantly in knowledge-intensive domains, their specific strengths vary. \textbf{Claude-Opus-4.6-thinking} sets the benchmark in \textit{Law} (65.54\%) and \textit{Humanities} (83.02\%), whereas \textbf{GPT-5.4-high} demonstrates a commanding lead in \textit{Finance} (84.65\%), outperforming competitive models like Doubao-2.0-pro (73.82\%). Notably, \textbf{GPT-5.4-high} demonstrates superior proficiency in the \textit{Education} domain (59.29\%), while \textbf{Claude-Opus-4.6-thinking} leads in \textit{STEM} (49.58\%), suggesting these latest thinking-augmented and scaled models offer stronger alignments with complex formal logic.

\begin{table}[t]
\centering
\caption{Model Performance Breakdown on the XpertBench-Gold subset (N=245) across Different Domains. The highest score in each category is highlighted in bold. EAS stands for Engineering and Applied Sciences; HSS stands for Humanities and Social Sciences.}
\label{tab:domain_breakdown}
\renewcommand{\arraystretch}{1.2}
\resizebox{\textwidth}{!}{ % Ensure the table fits within the text width
\begin{tabular}{lcccccc}
\toprule
\textbf{Model} & \textbf{Overall} & \textbf{Finance} & \textbf{Law} & \textbf{Education} & \textbf{EAS} & \textbf{HSS} \\
 % & ($N=245$) & ($N=81$) & ($N=51$) & ($N=30$) & ($N=51$) & ($N=33$) \\
\midrule
Claude-Opus-4.6-thinking & \textbf{66.20\%} & 73.25\% & \textbf{65.54\%} & 57.96\% & \textbf{49.58\%} & \textbf{83.02\%} \\
GPT-5.4-high & 64.78\% & \textbf{84.65\%} & 64.79\% & \textbf{59.29\%} & 42.84\% & 80.58\% \\
Doubao-2.0-pro & 64.51\% & 73.82\% & 65.06\% & 55.35\% & 44.88\% & 80.09\% \\
GPT-5.2-high & 61.03\% & 71.22\% & 56.88\% & 58.43\% & 46.13\% & 76.15\% \\
GPT-5-high & 58.70\% & 64.50\% & 54.70\% & 56.90\% & 48.20\% & 68.50\% \\
Gemini-3.1-pro & 55.05\% & 62.43\% & 53.88\% & 52.43\% & 37.30\% & 69.04\% \\
Gemini-3-pro & 53.43\% & 62.96\% & 48.59\% & 48.59\% & 36.34\% & 71.45\% \\
Kimi-k2-thinking & 53.30\% & 55.00\% & 58.00\% & 54.40\% & 37.10\% & 66.40\% \\
Gemini-3-flash & 50.98\% & 57.45\% & 50.22\% & 47.10\% & 31.56\% & 70.66\% \\
Claude-Sonnet-4.5-thinking-azure & 46.00\% & 44.50\% & 58.70\% & 44.50\% & 27.50\% & 54.90\% \\
DeepSeek-V3.1-Terminus-thinking & 45.10\% & 46.10\% & 51.90\% & 41.90\% & 32.20\% & 55.20\% \\
GLM-4.6 & 44.30\% & 52.70\% & 47.20\% & 39.10\% & 27.40\% & 49.70\% \\
Doubao-1.6-thinking & 42.10\% & 40.80\% & 49.40\% & 45.90\% & 29.50\% & 50.60\% \\
Gemini-2.5-Pro & 38.20\% & 30.30\% & 47.30\% & 47.90\% & 25.50\% & 52.30\% \\
\bottomrule
\end{tabular}
}
\end{table}

Furthermore, we observe a divergence between generation and logic capabilities in the \textit{STEM} category. Despite being the newer iteration, GPT-5.2-high (46.13\%) is slightly outperformed by \textbf{GPT-5-high} (48.20\%) in STEM tasks, indicating that the latter may retain an edge in strict calculation or formal logic consistency. Additionally, the domestic model \textbf{Kimi-k2-thinking} shows remarkable competitiveness, securing a top-tier position (53.30\%) that effectively matches Gemini-3-pro (53.43\%), particularly distinguishing itself with strong performance in legal contexts (58.00\%) where it rivals the top international models.

\textbf{Significant Capability Gaps Revealed at the Expert-Level Frontier.} Our benchmark proves to be exceptionally challenging for the current generation of LLMs, including both leading proprietary and open-source models. We observe a substantial performance delta: the state-of-the-art (SOTA) model (e.g., Claude-Opus-4.6-thinking and GPT-5.4-high), augmented with retrieval/search capabilities, achieve only a 
$\sim$65--66\% success rate. This stands in stark contrast to the majority of other models, which cluster around a $\sim$50\% completion rate. This wide performance gap underscores that current models lack the robust reasoning, planning, and knowledge synthesis required for genuine expert-level problem-solving, highlighting a significant frontier for future research.

\textbf{Pronounced Domain-Specific Specialization and Capability Divergence.} We find that ``expert intelligence'' is not monolithic. Our fine-grained evaluation reveals pronounced specialization among SOTA models, indicating that no single model universally dominates all domains. For instance, GPT-5.4-high demonstrates overwhelming superiority in \textit{Finance} (84.65\%), outperforming the runner-up by over 10 percentage points, yet it falls significantly behind the leading tier in \textit{STEM} (42.84\%). Conversely, Claude-Opus-4.6-thinking exhibits a more balanced "generalist" profile but still shows variance. These systematic specialization patterns imply that a single "best" model does not exist for expert workloads; instead, model choice must be matched to domain and task requirements.

\clearpage
\section{Contributions}

\noindent \textbf{Core Contributors} ($\alpha$-$\beta$ order) \\
Xue Liu, Xin Ma, Yuxin Ma, Yongchang Peng, Duo Wang$^\dagger$, Zhoufutu Wen, Ge Zhang$^\dagger$, Kaiyuan Zhang\\
(\texttt{\{wangduo.neuron, zhangge.eli\}@bytedance.com})

\vspace{1em}
\noindent \textbf{Contributors} ($\alpha$-$\beta$ order) \\
Xinyu Chen, Yida Ding, Tianci He, Jiani Hou, Liang Hu, Ziyun Huang, Yongzhe Hui, Jianpeng Jiao, Chennan Ju, Yingru Kong, Yiran Li, Jiashuo Liu, Mengyun Liu, Luyao Ma, Fei Ni, Yiqing Ni, Pengbo Niu, Yueyan Qiu, Yanle Ren, Xinyu Shen, Zilin Shi, Zaiyuan Wang, Wenjie Yue, Chun Zhang, Shiyu Zhang, Xinyi Zhang, Kaiwen Zhao, Zhenwei Zhu

\vspace{1em}
\noindent \textbf{Advisors} \\
Shanshan Wu (\url{wushanshan.sarah@bytedance.com})\\
Qi Zhao (\url{zhaoqi@bytedance.com})\\
Wenhao Huang (\url{huang.wenhao@bytedance.com}) \\

\vspace{2em}
\noindent $^\dagger$ denotes corresponding authors. \\
Contributors without explicit affiliations are from ByteDance Seed. During the work, Yuxin, Xinyu, Yida, Ziyun, Yongzhe, Chennan, Yingru, Yiran, Luyao, Yiqing, Pengbo, Yueyan, Yanle, Xinyu, Zilin, Wenjie, and Shiyu are interns at ByteDance Seed.

\section{Xpert Platform}

\subsection{What is Xpert Platform}
Xpert is an expert-level data service platform under ByteDance, committed to becoming the industry's leading specialized training data and evaluation solution provider. Our vision is to transform the deep knowledge and rich experience of experts across various industries into high-quality data, providing critical momentum for AGI and unlocking greater commercial and social value. The platform brings together approximately 3,000 rigorously selected experts, including master's and doctoral scholars from China's top-tier 985/211 universities as well as industry professionals with 2-10 years of rich practical experience in finance, law, healthcare, education, and others. Link:\url{https://xpert.bytedance.com/}

\subsection{Xpert Leaderboard Intro}
Unlike mainstream exam-oriented evaluations, Xpert Leaderboard focuses on assessing AI's ability to solve expert-level complex tasks in the real world, dedicated to driving AI to create greater economic value. Link:\url{https://xpert.bytedance.com/leaderboard}

\clearpage

\bibliographystyle{unsrt}
\bibliography{main}

@article{medqa,
  title={What Disease does this Patient Have? A Large-scale Open Domain Question Answering Dataset from Medical Exams},
  author={Jin, Di and Pan, Eileen and Oufattole, Nassim and Weng, Wei-Hung and Fang, Hanyi and Szolovits, Peter},
  journal={Applied Sciences},
  volume={11},
  number={14},
  pages={6421},
  year={2021}
}

@inproceedings{finben,
  title={{FinBen}: A Holistic Financial Benchmark for Large Language Models},
  author={Xie, Qianqian and Han, Weiguang and Chen, Zhengyu and Xia, Ruoyu and
          Zhang, Xiao and He, Yueru and Xiao, Mengxi and Li, Dong and
          Dai, Yongfu and Feng, Duanyu and others},
  booktitle={Advances in Neural Information Processing Systems},
  volume={37},
  year={2024}
}

@article{humanitylastexam,
  title={Humanity's Last Exam},
  author={Phan, Long and Gatti, Alice and Han, Ziwen and Li, Nathaniel and
          Hu, Josephina and Zhang, Hugh and Shi, Sean and Choi, Michael and
          Agrawal, Anish and Chopra, Arnav and others},
  journal={Nature},
  year={2026},
  publisher={Springer Nature}
}

@article{frontiermath,
  title={{FrontierMath}: A Benchmark for Evaluating Advanced Mathematical
         Reasoning in {AI}},
  author={Glazer, Elliot and Erdil, Ege and Besiroglu, Tamay and
          Chicharro, Diego and Chen, Evan and Gunning, Alex and
          Olsson, Caroline Falkman and Denain, Jean-Stanislas and
          Ho, Anson and Santos, Emily de Oliveira and others},
  journal={arXiv preprint arXiv:2411.04872},
  year={2024}
}

@article{browsecomp,
  title={{BrowseComp}: A Simple Yet Challenging Benchmark for Browsing Agents},
  author={Wei, Jason and Cho, Mia and Cummings, Aidan and Guo, Karina and
          Hu, Shixiang Shane and Kang, Simon and Khlaaf, Heidy and Miao, Neal and
          Neyman, Oam and Rubin, Noa and others},
  journal={arXiv preprint arXiv:2501.12959},
  year={2025}
}

@inproceedings{deepresearchbench,
  title={{DeepResearch Bench}: A Comprehensive Benchmark for Deep Research Agents},
  author={Du, Mingxuan and Xu, Benfeng and Zhu, Chiwei and Zhang, Licheng and
          Wang, Xiaorui and Mao, Zhendong},
  booktitle={International Conference on Learning Representations},
  year={2026}
}

@article{deer,
  title={{DEER}: A Comprehensive and Reliable Benchmark for Deep Research
         Agents on Expert-Level Research Tasks},
  author={Zhang, Yifan and Chen, Yifan and Liu, Haoyang and
          Fang, Zhicheng and others},
  journal={arXiv preprint arXiv:2512.17776},
  year={2025}
}

@inproceedings{wildbench,
  title={{WildBench}: Benchmarking {LLMs} with Challenging Tasks from Real
         Users in the Wild},
  author={Lin, Bill Yuchen and Deng, Yuntian and Chandu, Khyathi Raghavi and
          Brahman, Faeze and Srivastava, Abhilasha and Ravichander, Abhilasha and
          Choi, Yejin and Smith, Noah A.},
  booktitle={Advances in Neural Information Processing Systems},
  volume={37},
  year={2024}
}

@inproceedings{judgebench,
  title={{JudgeBench}: A Benchmark for Evaluating {LLM}-based Judges},
  author={Zhou, Sijun and Huang, Nuo and Xu, Ran and Yan, Renren and
          Li, Muning and Xiao, Yanghua and Hemphill, Libby},
  booktitle={International Conference on Learning Representations},
  year={2025}
}

@article{rubriceval,
  title={{RubricEval}: A Scalable Human-{LLM} Evaluation Framework for
         Open-Ended Tasks},
  author={Bhat, Meera and Fang, Xi and Steinhardt, Jacob},
  journal={Stanford CS224N Final Reports},
  year={2024}
}

@inproceedings{mmlupro,
  title={{MMLU}-Pro: A More Robust and Challenging Multi-Task Language Understanding Benchmark},
  author={Wang, Yubo and Ma, Xueguang and Zhang, Ge and Ni, Yuansheng and Chandra, Abhranil and Guo, Shiguang and Ren, Weiming and Arulraj, Aaran and He, Xuan and Jiang, Ziyan and others},
  booktitle={Advances in Neural Information Processing Systems},
  volume={37},
  year={2024}
}

@inproceedings{gpqa,
  title={{GPQA}: A Graduate-Level {Google}-Proof {Q\&A} Benchmark},
  author={Rein, David and Hou, Betty Li and Stickland, Asa Cooper and Petty, Jackson and Pang, Richard Yuanzhe and Dirani, Julien and Michael, Julian and Bowman, Samuel R.},
  booktitle={Proceedings of the First Conference on Language Modeling},
  year={2024},
  eprint={2311.12022},
  archivePrefix={arXiv}
}

@misc{wang2023large,
      title={Large Language Models are not Fair Evaluators}, 
      author={Peiyi Wang and Lei Li and Liang Chen and Zefan Cai and Dawei Zhu and Binghuai Lin and Yunbo Cao and Qi Liu and Tianyu Liu and Zhifang Sui},
      year={2023},
      eprint={2305.17926},
      archivePrefix={arXiv},
      primaryClass={cs.CL}
}

@inproceedings{pubmedqa,
  title={{PubMedQA}: A Dataset for Biomedical Research Question Answering},
  author={Jin, Qiao and Dhingra, Bhuwan and Liu, Zhengping and Cohen, William and Lu, Xinghua},
  booktitle={Proceedings of the 2019 Conference on Empirical Methods in Natural Language Processing and the 9th International Joint Conference on Natural Language Processing (EMNLP-IJCNLP)},
  pages={2567--2577},
  year={2019}
}

@inproceedings{scibench,
  title={{SciBench}: Evaluating College-Level Scientific Problem-Solving Abilities of Large Language Models},
  author={Wang, Xiaoxuan and Hu, Ziniu and Lu, Pan and Zhu, Yanqiao and Zhang, Jieyu and Subramaniam, Satyen and Loomba, Arjun R. and Zhang, Shichang and Sun, Yizhou and Wang, Wei},
  booktitle={Proceedings of the Forty-First International Conference on Machine Learning},
  year={2024}
}

@inproceedings{legalbench,
  title={{LegalBench}: A Collaboratively Built Benchmark for Measuring Legal Reasoning in Large Language Models},
  author={Guha, Neel and Nyarko, Julian and Ho, Daniel E. and R{\'e}, Christopher and Chilton, Adam and Narayana, Aditya and Chohlas-Wood, Alex and Peters, Austin and Waldon, Brandon and Rockmore, Daniel N. and others},
  booktitle={Advances in Neural Information Processing Systems},
  volume={36},
  pages={59201--59242},
  year={2023},
  publisher={Curran Associates, Inc.}
}

@inproceedings{agentbench,
  title={{AgentBench}: Evaluating {LLMs} as Agents},
  author={Liu, Xiao and Yu, Hao and Zhang, Hanchen and Xu, Yifan and Lei, Xuanyu and Lai, Hanyu and Gu, Yu and Ding, Hangliang and Men, Kaiwen and Yang, Kejuan and others},
  booktitle={The Twelfth International Conference on Learning Representations},
  year={2024}
}

@inproceedings{gaia,
  title={{GAIA}: a benchmark for General {AI} Assistants},
  author={Mialon, Gr{\'e}goire and Fourrier, Cl{\'e}mentine and Swift, Craig and Wolf, Thomas and LeCun, Yann and Scialom, Thomas},
  booktitle={The Twelfth International Conference on Learning Representations},
  year={2024},
  eprint={2311.12983},
  archivePrefix={arXiv}
}

@article{helm,
  title={Holistic Evaluation of Language Models},
  author={Liang, Percy and Bommasani, Rishi and Lee, Tony and Tsipras, Dimitris and Soylu, Dilara and Yasunaga, Michihiro and Zhang, Yian and Narayanan, Deepak and Wu, Yuhuai and Kumar, Ananya and others},
  journal={Transactions on Machine Learning Research},
  year={2023}
}

@inproceedings{chatbotarena,
  title={Chatbot Arena: An Open Platform for Evaluating {LLMs} by Human Preference},
  author={Chiang, Wei-Lin and Zheng, Lianmin and Sheng, Ying and Angelopoulos, Anastasios Nikolas and Li, Tianle and Li, Dacheng and Zhang, Hao and Zhu, Banghua and Jordan, Michael and Gonzalez, Joseph E. and Stoica, Ion},
  booktitle={Proceedings of the Forty-First International Conference on Machine Learning},
  year={2024}
}

@misc{simpleqa,
  title={Measuring Short-Form Factuality in Large Language Models},
  author={Wei, Jason and Nguyen, Karina and Chung, Hyung Won and Jiao, Yunxin Joy and Papay, Spencer and Glaese, Amelia and Schulman, John and Fedus, William},
  year={2024},
  eprint={2411.04368},
  archivePrefix={arXiv},
  primaryClass={cs.CL}
}

@misc{alpacaeval,
  title = {{{AlpacaEval: An Automatic Evaluator for Instruction-following Models}}},
  author = {Li, Xuechen and Zhang, Tianyi and Dubois, Yann and others},
  year = {2023},
  publisher = {GitHub},
  howpublished = {\url{https://github.com/tatsu-lab/alpaca_eval}}
}

@inproceedings{mtbench,
  title={Judging {LLM}-as-a-Judge with {MT-Bench} and Chatbot Arena},
  author={Zheng, Lianmin and Chiang, Wei-Lin and Sheng, Ying and Zhuang, Siyuan and Wu, Zhanghao and Zhuang, Yonghao and Lin, Zi and Li, Zhuohan and Li, Dacheng and Xing, Eric P. and Zhang, Hao and Gonzalez, Joseph E. and Stoica, Ion},
  booktitle={Advances in Neural Information Processing Systems},
  volume={36},
  year={2023}
}

@inproceedings{bi2024prometheus,
  title={Prometheus: Inducing Fine-grained Evaluation Capability in Language Models},
  author={Bi, Seungone and Koch, Christoph and Chen, Guoyin and Agarwal, Trevor and others},
  booktitle={The Twelfth International Conference on Learning Representations},
  year={2024}
}

@misc{arenahard,
  title = {{{Arena-Hard Auto: Evaluating LLMs with Human-in-the-loop Standards}}},
  author = {Li, Tianle and Chiang, Wei-Lin and Frick, Evan and others},
  year = {2024},
  publisher = {LMSYS Org},
  howpublished = {\url{https://lmsys.org/blog/2024-04-19-arena-hard/}}
}

\clearpage

\beginappendix
\appendix
\counterwithin{figure}{section}
\setcounter{figure}{0}
\counterwithin{table}{section}
\setcounter{table}{0}

\section{Example Tasks and Scoring Rubrics}
\label{sec:appendix}

In this appendix, we present representative example tasks and their corresponding scoring rubrics for each of the five evaluation categories: Finance, Law, Education, STEM, and Humanities \& Social Sciences (HSS).

%% ============================================================
%% A. Finance
%% ============================================================
\subsection{Finance}
\label{sec:appendix-finance}

\noindent\textit{Corporate strategy analysis, financial reporting, market research, and business case studies.}

\subsubsection{Example Task}

As a senior analyst at a major credit rating agency, your task is to prepare an in-depth comparative analysis for the rating committee, evaluating the operational performance and financial discipline of two U.S.\ aerospace and defense giants---Lockheed Martin and Northrop Grumman---against the macro backdrop of heightened global geopolitical tensions and sustained defense budget growth during 2022--2023.

Conduct precise quantitative comparisons and analyses of the following companies strictly within the complete macroeconomic cycle period from January 1, 2022, to December 31, 2023:

\begin{enumerate}
    \item Lockheed Martin Corporation (NYSE: LMT) -- The world's largest defense contractor, renowned for its dominant position in aviation (e.g., the F-35 fighter jet).
    \item Northrop Grumman Corporation (NYSE: NOC) -- A defense giant with formidable technological moats in aerospace, mission systems, and strategic weapons (e.g., the B-21 stealth bomber).
\end{enumerate}

\noindent\textbf{Core Analysis Requirements:}

\begin{enumerate}
    \item \textbf{Future Revenue Visibility Comparison:}
    \begin{itemize}
        \item The ``Book-to-Bill Ratio'' serves as the lifeline for gauging a defense company's future revenue growth potential. Locate the ``Net Sales'' for FY2023 in both companies' annual reports. Additionally, find the ``Net Orders'' or calculate it from the change in ``Total Backlog'' within the Management Discussion and Analysis (MD\&A) or Financial Data Summary sections of their annual reports.
        \item Based on the above data, calculate each company's ``Order-to-Sales Ratio'' for FY 2023. What does a ratio above or below 1.0 signify? How does the difference in their order-to-sales ratios reflect their varying performance in securing new contracts and replenishing order backlogs?
    \end{itemize}

    \item \textbf{Core Profitability Engine Comparison:}
    \begin{itemize}
        \item A company's profit engine determines its earnings quality. Examine the ``Business Segment Information'' in the notes to the financial statements for both companies. Identify and compare the ``Operating Margin'' for all business segments (e.g., Aerospace, Missiles, Mission Systems, Space) in FY2023.
        \item Identify the core business segment with the highest operating margin for each company in FY2023. How do the differences between these high-margin segments reveal fundamental distinctions in technological barriers, market positioning, and profitability between the two companies?
    \end{itemize}

    \item \textbf{Cash Flow Generation Efficiency Comparison:}
    \begin{itemize}
        \item The ability to convert accounting profits into actual cash is crucial for assessing a company's financial health. Locate the following items in each company's consolidated financial statements for fiscal year 2023: ``Net cash provided by operating activities,'' ``Capital expenditures,'' and ``Net earnings.''
        \item Based on these figures, calculate each company's ``Free Cash Flow Conversion Rate'' for FY 2023. How does a significant disparity in this ratio reveal differences in their efficiency regarding working capital management, project execution, and converting profits into available cash?
    \end{itemize}

    \item \textbf{Conclusion Summary:}
    \begin{itemize}
        \item Based on the quantitative analysis of future revenue visibility, core profit engines, and cash flow efficiency, summarize the key factors (e.g., ability to win new contracts, technological moats in high-margin businesses, and cash flow management efficiency) that led to the significant divergence in operational and financial performance between Lockheed Martin and Northrop Grumman during FY2023---a year of increased defense budgets.
    \end{itemize}
\end{enumerate}

Ensure your analysis is grounded in verifiable data and logically rigorous. Avoid any future predictions or investment advice.

\subsubsection{Scoring Rubric}

{\small
\begin{longtable}{p{9cm}>{\centering\arraybackslash}p{1.2cm}>{\centering\arraybackslash}p{2cm}>{\centering\arraybackslash}p{2.8cm}}
\caption{Scoring rubric for the Finance example task.}
\label{tab:rubric-finance} \\
\toprule
\multicolumn{1}{c}{\textbf{Criterion}} & \textbf{Points} & \textbf{Importance} & \textbf{Tag} \\
\midrule
\endfirsthead

\multicolumn{4}{l}{\small\textit{Table~\ref{tab:rubric-finance} (continued)}} \\
\toprule
\multicolumn{1}{c}{\textbf{Criterion}} & \textbf{Points} & \textbf{Importance} & \textbf{Tag} \\
\midrule
\endhead

\bottomrule
\endfoot
Based on FY2023 data, calculate Lockheed Martin's book-to-bill ratio as approximately 1.16. The calculation must use correct net order amounts and total revenue. A $\pm$2\% margin of error is allowed. & 10 & Essential & Authenticity \\
\midrule
Based on FY2023 data, calculate Northrop Grumman's book-to-bill ratio as approximately 1.14. The calculation must use correct net order amounts and total revenue. A $\pm$2\% margin of error is allowed. & 10 & Essential & Authenticity \\
\midrule
Accurately explain the meaning of a book-to-bill ratio (Book-to-Bill) above 1.0, i.e., new orders exceed current period revenue, indicating strong future revenue visibility. & 10 & Essential & Depth \\
\midrule
Accurately identify LMT's business segment with the highest operating margin in 2023 as ``Missiles and Fire Control.'' & 10 & Essential & Authenticity \\
\midrule
Accurately state that the ``Missiles and Fire Control'' segment's operating margin is approximately 13.7\%. A $\pm$0.3\% margin of error is allowed. & 10 & Essential & Authenticity \\
\midrule
Accurately identify Northrop Grumman's business segment with the highest operating margin in 2023 as ``Mission Systems.'' & 10 & Essential & Authenticity \\
\midrule
Accurately state that the ``Mission Systems'' segment's operating margin is approximately 14.8\%. A $\pm$0.3\% margin of error is allowed. & 10 & Essential & Authenticity \\
\midrule
Compare the differences in high-margin business segments between the two companies and relate them to their respective technological barriers or market positioning (e.g., precision strike systems vs.\ advanced electronic systems). & 7 & Important & Depth \\
\midrule
Based on FY2023 data, calculate Lockheed Martin's free cash flow conversion rate as approximately 90.0\%. A $\pm$2\% margin of error is allowed. & 7 & Important & Authenticity \\
\midrule
Based on FY2023 data, calculate Northrop Grumman's free cash flow conversion rate as approximately 102.1\%. A $\pm$2\% margin of error is allowed. & 7 & Important & Authenticity \\
\midrule
Accurately point out that Northrop Grumman has higher cash flow efficiency and relate it to superior working capital management capabilities. & 7 & Important & Depth \\
\midrule
At the end of the report, synthesize the three quantitative analyses to summarize the core drivers behind performance differences between the two companies. & 7 & Important & Instruction Following \\
\bottomrule
\end{longtable}
}

%% ============================================================
%% B. Law
%% ============================================================
\subsection{Law}
\label{sec:appendix-law}

\noindent\textit{Legal document drafting, complex legal reasoning, evidence analysis, and legal argumentation.}

\subsubsection{Example Task}

I am the Legal Affairs Manager of a state-owned enterprise supply chain company in Guangxi Province (hereinafter referred to as ``Guangxi Company''). You are aware of the legal facts as shown in Attachment 1.

As the Legal Affairs Manager of Guangxi Company, based on the aforementioned legal facts and in conjunction with legal practice and theory:

\begin{enumerate}
    \item Based on legal theory, determine whether the agreement signed between Guangxi Company and China Construction Bank on June 1, 2023, constitutes a loan relationship or a factoring contract relationship?
    \item What is the validity of the Factoring Financing Agreement signed between a financing guarantee company in Yunnan Province and Guangxi Company?
    \item If the financing guarantee company in Yunnan Province asserts rights against Guangxi Company or Yunnan Company based on the factoring contract relationship under the Factoring Financing Agreement, how should the liability be allocated between Yunnan Company and Guangxi Company?
\end{enumerate}

\subsubsection{Scoring Rubric}

{\small
\begin{longtable}{p{9cm}>{\centering\arraybackslash}p{1.2cm}>{\centering\arraybackslash}p{2cm}>{\centering\arraybackslash}p{2.8cm}}
\caption{Scoring rubric for the Law example task.}
\label{tab:rubric-law} \\
\toprule
\multicolumn{1}{c}{\textbf{Criterion}} & \textbf{Points} & \textbf{Importance} & \textbf{Tag} \\
\midrule
\endfirsthead

\multicolumn{4}{l}{\small\textit{Table~\ref{tab:rubric-law} (continued)}} \\
\toprule
\multicolumn{1}{c}{\textbf{Criterion}} & \textbf{Points} & \textbf{Importance} & \textbf{Tag} \\
\midrule
\endhead

\bottomrule
\endfoot
Accurately identify the legal relationship between Guangxi Company and China Construction Bank as accounts receivable pledge financing, rather than a factoring contract relationship. & 7 & Important & Requirement Identification \\
\midrule
Ability to clarify the core reasons for determining a loan contract relationship, i.e., the financing behavior in this case lacks the core element of ``accounts receivable transfer'' required for a factoring contract. & 7 & Important & Logicality \\
\midrule
Accurately analyze that lack of factoring qualifications does not necessarily render the Factoring Financing Agreement invalid, and reach a conclusion favoring contract validity. & 7 & Important & Professionalism \\
\midrule
Ability to cite key legal principles or provisions to argue why lack of factoring qualifications does not constitute an ``effectiveness-mandatory provision'' that would invalidate the contract. & 7 & Important & Professionalism \\
\midrule
Accurately point out that in recourse factoring, the factor (financing guarantee company) has no sequential restrictions when asserting rights against the creditor (Guangxi Company) and the debtor (Yunnan Company). & 7 & Important & Accuracy \\
\midrule
Ability to analyze the legal consequences and liability allocation for all parties if the Factoring Financing Agreement is deemed invalid. & 7 & Important & Logicality \\
\midrule
Ability to cite Article 761 of the Civil Code regarding the definition of factoring contracts as the legal basis for analysis. & 7 & Important & Accuracy \\
\midrule
Ability to recognize and elaborate on judicial practice disputes regarding the impact of operating factoring business without qualifications on contract validity. & 7 & Important & Professionalism \\
\midrule
Ability to elaborate on theoretical viewpoints supporting ``no sequential restrictions on recourse,'' such as the ``assignment guarantee theory.'' & 7 & Important & Professionalism \\
\midrule
Ability to identify and point out that the subject matter of the Factoring Financing Agreement in this case is ``future accounts receivable'' and recognize its transferability. & 7 & Important & Requirement Identification \\
\midrule
Must not mistakenly equate accounts receivable ``pledge'' with ``transfer,'' thereby misjudging a loan relationship as a factoring relationship. & 7 & Important & Accuracy \\
\midrule
Must not absolutize legal application; must not simply deem the Factoring Financing Agreement invalid based on ``lack of qualifications'' or ``exceeding business scope'' without conducting in-depth analysis. & 7 & Important & Logicality \\
\midrule
Must not misunderstand recourse factoring; must not erroneously believe the factor must first exhaust remedies against the accounts receivable debtor (Yunnan Company) before seeking recourse from the creditor (Guangxi Company). & 7 & Important & Accuracy \\
\bottomrule
\end{longtable}
}

%% ============================================================
%% C. Education
%% ============================================================
\subsection{Education}
\label{sec:appendix-education}

\noindent\textit{Instructional \& curriculum design, pedagogical assessment, student learning analysis, and educational research synthesis.}

\subsubsection{Example Task}

Please create a philosophical dialogue script about Confucius and Socrates from the perspective of a screenwriter.

Set the encounter between the two sages in their twilight years, disregarding temporal and spatial constraints. The script's core theme revolves around a profound debate concerning views on life and death. The dialogue style should align with the linguistic characteristics reflected in the works of Confucius and Socrates. The dialogue must accurately reflect their respective philosophical beliefs and views on life and death, incorporating relevant biographical facts from their lives to support their arguments. The entire script must be written in modern Chinese. Reasonable artistic interpretation is permitted, but historical accuracy and philosophical integrity must be maintained. The final output must adhere to standard screenplay formatting requirements.

\subsubsection{Scoring Rubric}

{\small
\begin{longtable}{p{9cm}>{\centering\arraybackslash}p{1.2cm}>{\centering\arraybackslash}p{2cm}>{\centering\arraybackslash}p{2.8cm}}
\caption{Scoring rubric for the Education example task.}
\label{tab:rubric-education} \\
\toprule
\multicolumn{1}{c}{\textbf{Criterion}} & \textbf{Points} & \textbf{Importance} & \textbf{Tag} \\
\midrule
\endfirsthead

\multicolumn{4}{l}{\small\textit{Table~\ref{tab:rubric-education} (continued)}} \\
\toprule
\multicolumn{1}{c}{\textbf{Criterion}} & \textbf{Points} & \textbf{Importance} & \textbf{Tag} \\
\midrule
\endhead

\bottomrule
\endfoot
Teaching objectives adopt ``three-dimensional objectives'' with point-by-point elaboration, and each dimension must list at least one observable, measurable specific behavioral objective. & 7 & Important & Standardization \\
\midrule
Each dimension of teaching objectives requires layered design, with clear evaluation criteria set for three categories of students: better, moderate, and more severe. & 7 & Important & Completeness \\
\midrule
Each dimension of teaching objectives must use measurable action verbs and provide specific conditions or standards (e.g., ``able to read correctly within 3 attempts''). & 7 & Important & Standardization \\
\midrule
Teaching focus must clearly state ``able to write time format correctly (HH:MM)'' and provide one example. & 7 & Important & Instruction Following \\
\midrule
Difficulty explanation should simultaneously provide the main methods to overcome difficulties and corresponding teaching aids. & 10 & Essential & Completeness \\
\midrule
All listed teaching methods must appear in at least two specific application scenarios in the subsequent teaching process. & 7 & Important & Completeness \\
\midrule
Teaching process includes five stages: introduction, new content, consolidation, summary, and homework assignment, with approximate time allocation noted for each stage. & 7 & Important & Instruction Following \\
\midrule
Each stage of instructional design must reflect student-centeredness and list at least two opportunities for student autonomous inquiry or self-evaluation. & 7 & Important & Comprehensive Application \\
\midrule
Introduction stage must review analog clock knowledge and compare with digital clocks through teacher demonstration or physical demonstration. & 10 & Essential & Instruction Following \\
\midrule
Instructional design needs to incorporate specific vocational life cases. & 10 & Essential & Completeness \\
\midrule
New content section must use mnemonics to teach reading digital clocks and immediately have students complete one practice using the mnemonic after explanation. & 10 & Essential & Instruction Following \\
\midrule
New content section must mention using Seewo games or other interactive software for immediate consolidation. & 10 & Essential & Completeness \\
\midrule
New content section must provide at least two complete dialogue examples between teacher--student or student--student, reflecting interaction details. & 10 & Essential & Completeness \\
\midrule
Consolidation/application stage reflects participation of all students, with each activity accompanied by detailed rule descriptions. & 7 & Important & Completeness \\
\midrule
Homework assignment must be layered and life-oriented, providing one example (such as a sample time record table). & 7 & Important & Completeness \\
\midrule
Identify and note: Currently, the People's Education Press compulsory education textbooks in mainland China do not include clock-related knowledge at the elementary level; therefore, other materials will be referenced. & 7 & Important & Requirement Identification \\
\bottomrule
\end{longtable}
}

%% ============================================================
%% D. STEM
%% ============================================================
\subsection{STEM}
\label{sec:appendix-stem}

\noindent\textit{Technical problem solving, experimental design, data analysis, and scientific reasoning.}

\subsubsection{Example Task}

A research group needs to construct a genomic cosmid library using pJTU++++ (approximately 9.5 kb in size) as the vector. After completing the cosmid library construction, the empty vector rate must be verified. Twenty single colonies (labeled 1--20) were randomly selected. Plasmids were extracted from these colonies and transformed into \textit{E.\ coli} DH5$\alpha$. Verification was performed using restriction enzyme \textit{Mbo}I digestion. Electrophoresis results are shown in Figure 1 of the second image. Lanes 1--20 display digested cosmids, while lanes 21--22 show undigested cosmids from columns 1 and 2. These plasmids were then transferred into \textit{E.\ coli} ET12567. After re-extracting the plasmids, restriction enzyme digestion was performed using \textit{Mbo}I and its homolog \textit{Sau}3AI (both share identical cleavage sites, as shown in the first image below). The electrophoresis results are depicted in Figure 2 of the second image.

\noindent\textbf{Questions:}
\begin{enumerate}
    \item Based on the above conditions and comprehensive analysis of electrophoresis patterns in Figures 1 and 2 of the second image, what is the empty vector rate among the 20 selected single colonies?
    \item What is the minimum average size of the foreign fragment in the 20 plasmids shown in the figure?
    \item Why do Figures 1 and 2 in the second figure show differences when the same plasmid is digested with the same restriction enzyme \textit{Mbo}I and its restriction enzyme \textit{Sau}3AI?
\end{enumerate}

\noindent\textbf{Attachments:}

% Placeholder for STEM attachment images
\begin{figure}[h]
    \centering
    \includegraphics[width=0.4\textwidth]{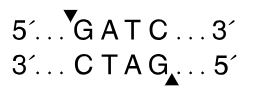} % 建议把这个图调小一点
    \caption{The recognition sequence and cleavage sites (indicated by triangles) of the restriction enzymes \textit{Mbo}I and \textit{Sau}3AI. Both enzymes recognize the same $5^\prime$-GATC-$3^\prime$ nucleotide sequence.}
    \label{fig:stem-restriction-site}
\end{figure}

\begin{figure}[H]
    \centering
    \includegraphics[width=0.5\textwidth]{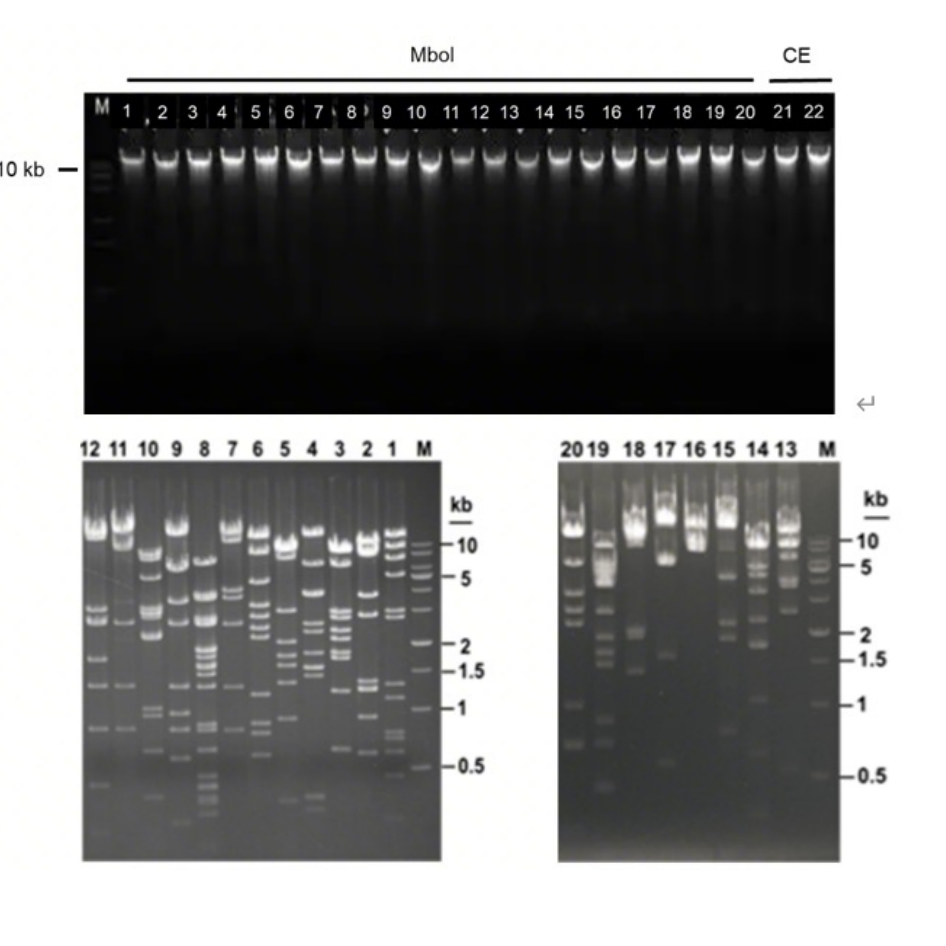}
    \caption{Agarose gel electrophoresis results for plasmid verification. \textbf{Top (Figure 1):} Digestion profiles of plasmids extracted from \textit{E.\ coli} DH5$\alpha$. Lanes 1--20 are samples treated with \textit{Mbo}I; lanes 21--22 (CE) are undigested control plasmids. \textbf{Bottom (Figure 2):} Digestion profiles of the same plasmids after being transferred to and extracted from \textit{E.\ coli} ET12567. The left panel shows digestion with \textit{Mbo}I, and the right panel shows digestion with \textit{Sau}3AI. 'M' denotes the DNA molecular weight marker (ladder).}
    \label{fig:stem-gel-results}
\end{figure}

\subsubsection{Scoring Rubric}

{\small
\begin{longtable}{p{9cm}>{\centering\arraybackslash}p{1.2cm}>{\centering\arraybackslash}p{2cm}>{\centering\arraybackslash}p{2.8cm}}
\caption{Scoring rubric for the STEM example task (Part 1).}
\label{tab:rubric-stem-1} \\
\toprule
\multicolumn{1}{c}{\textbf{Criterion}} & \textbf{Points} & \textbf{Importance} & \textbf{Tag} \\
\midrule
\endfirsthead

\multicolumn{4}{l}{\small\textit{Table~\ref{tab:rubric-stem-1} (continued)}} \\
\toprule
\multicolumn{1}{c}{\textbf{Criterion}} & \textbf{Points} & \textbf{Importance} & \textbf{Tag} \\
\midrule
\endhead

\bottomrule
\endfoot
Analyze the reason for unsuccessful digestion: Dam methylation modification in DH5$\alpha$ blocks restriction enzyme digestion. & 10 & Essential & Information Processing \\
\midrule
Determine the state of fragments in the map: bands represent circular plasmids whose migration rate is affected by conformation (supercoiled, open circular, etc.); they cannot reflect actual size and cannot be used to determine empty vectors. & 10 & Essential & Professionalism \\
\midrule
Observe the restriction map and determine that plasmids have been successfully digested into linear fragments. The sum of bands can reflect the total length of ``vector $+$ foreign fragment,'' which is the key basis for determining empty vectors. & 10 & Essential & Professionalism \\
\midrule
Analyze the reason for successful digestion: ET12567 is a methylation-deficient strain (dam$^-$ dcm$^-$), and digestion successfully produces linear fragments. & 7 & Important & Professionalism \\
\midrule
Explain the difference in migration rate between circular plasmids (especially supercoiled) in electrophoresis and linear DNA used as markers (supercoiled migrates faster than linear DNA of the same size). & 7 & Important & Professionalism \\
\midrule
Point out that bands near 10 kb in Figure 1 only represent circular plasmid conformations, unrelated to the actual size of the plasmid (vector $+$ foreign fragment), and cannot directly determine whether it is empty based on position. & 10 & Essential & Information Processing \\
\midrule
Empty plasmids contain only the vector (9.5 kb); after digestion in ET12567, the sum of their linear fragments should be close to 9.5 kb (vector bands only). & 10 & Essential & Accuracy \\
\midrule
In Figure 2, the sum of restriction fragments of all lanes is far greater than 9.5 kb, proving that all contain foreign fragments, with an empty vector rate of 0. & 10 & Essential & Accuracy \\
\midrule
Explain that ``sum of fragments'' must include all visible bands (to avoid missing small fragments); if empty, it only includes vector bands. & 7 & Important & Accuracy \\
\midrule
Compare the differences in restriction maps between empty and non-empty vectors in ET12567 (empty: single or few vector bands; non-empty: multiple fragments, sum $>$ vector). & 10 & Essential & Professionalism \\
\midrule
Identify that Figure 2 has been successfully digested into linear fragments, and accurate size determination must rely on restriction fragments in Figure 2. & 10 & Essential & Accuracy \\
\midrule
Specifically explain the migration rate order of three conformations (supercoiled $>$ linear $>$ open circular). & 7 & Important & Accuracy \\
\midrule
Give examples to illustrate the migration position differences between circular and linear DNA of the same size (e.g., a supercoiled plasmid larger than 10 kb may migrate to the position of the 10 kb linear marker). & 7 & Important & Professionalism \\
\midrule
Calculation formula: foreign fragment size $=$ (sum of all linear fragment sizes in a lane in Figure 2) $-$ vector size (9.5 kb). & 10 & Essential & Professionalism \\
\midrule
Lower limit determination must consider band overlap (large fragment overlap leading to underestimation of sums) and small fragment omission (low brightness or running off the gel); the minimum fragment sum among the 20 samples yields a foreign fragment size $\geq$19 kb. & 7 & Important & Professionalism \\
\midrule
Give an example of fragment sum estimation for a lane (e.g., band sum $\approx$25 kb, foreign fragment $\approx$25$-$9.5$=$15.5 kb). & 7 & Important & Professionalism \\
\midrule
Explain that ``at least greater than 15 kb'' is a conservative lower limit (actual average approximately 20 kb), to avoid underestimation of true size due to technical errors. & 7 & Important & Accuracy \\
\midrule
Restriction-modification systems consist of methyltransferases (marking own DNA) and restriction endonucleases (cutting unmarked foreign DNA). & 10 & Essential & Accuracy \\
\midrule
Methyltransferases modify recognition sequences (e.g., Dam methylase modifies the N6 position of adenine in GATC), preventing own DNA from being cut by homologous restriction endonucleases. & 10 & Essential & Accuracy \\
\midrule
Distinguish between recognition sites of Dam (adenine methylation in GATC sequence) and Dcm (cytosine methylation in CCWGG sequence) methylases. & 7 & Important & Accuracy \\
\midrule
Explain the ``self-protection'' logic of restriction-modification systems (methylation marks as ``self,'' unmarked as ``foreign'' to be cut). & 10 & Essential & Professionalism \\
\midrule
DH5$\alpha$: dam$^+$ dcm$^+$, contains functional Dam and Dcm methylases; plasmid GATC sites are modified by Dam methylation. & 10 & Essential & Accuracy \\
\midrule
ET12567: dam$^-$13::Tn9 (Dam-deficient), dcm$^-$6 (Dcm-deficient); plasmid GATC sites have no methylation modification. & 10 & Essential & Accuracy \\
\midrule
Specifically explain the mechanism by which dam gene mutation (transposon insertion) in ET12567 causes Dam methylase inactivation. & 7 & Important & Professionalism \\
\midrule
Compare specific methylation modification sites in plasmids between the two strains: DH5$\alpha$: A methylation in GATC; ET12567: no such modification. & 10 & Essential & Professionalism \\
\midrule
\textit{Mbo}I: sensitive to Dam methylation; cannot cut if recognition site GATC is methylated (e.g., plasmids in DH5$\alpha$). & 10 & Essential & Accuracy \\
\midrule
\textit{Sau}3AI: an isoschizomer of \textit{Mbo}I (recognizes GATC), but insensitive to Dam methylation; can cut regardless of whether GATC is methylated. & 10 & Essential & Accuracy \\
\midrule
In Figure 1, DH5$\alpha$ plasmids cannot be cut by \textit{Mbo}I due to GATC methylation; in Figure 2, ET12567 plasmids have no methylation, so both \textit{Mbo}I and \textit{Sau}3AI can cut, producing linear fragments. & 10 & Essential & Accuracy \\
\midrule
Point out the methylation sensitivity specificity of \textit{Sau}3AI (only sensitive to CpG methylation, insensitive to Dam/Dcm methylation; in this case, since ET12567 has no methylation, both enzymes yield the same result). & 10 & Essential & Accuracy \\
\midrule
Explain the definition of ``isoschizomers'' (recognize the same site but may have different sensitivity to modifications) and compare the enzymatic property differences between \textit{Mbo}I and \textit{Sau}3AI. & 10 & Essential & Professionalism \\
\bottomrule
\end{longtable}
}

%% ============================================================
%% E. HSS (Humanities & Social Sciences)
%% ============================================================
\subsection{Humanities \& Social Sciences (HSS)}
\label{sec:appendix-hss}

\noindent\textit{Critical analysis, theory application, academic writing, and research methods.}

\subsubsection{Example Task}

In the capacity of a screenwriter, please create a philosophical dialogue script featuring Confucius and Socrates.

The premise is that the two sages meet in their final years, transcending temporal and spatial constraints. The script's core theme must be a profound debate centered on their respective philosophies of life and death.

The dialogue style must be consistent with the linguistic styles demonstrated by Confucius and Socrates in their respective works (e.g., the Analects and the Platonic dialogues). The dialogue content must accurately conform to their distinct philosophical ideas and views on mortality, and may moderately integrate biographical facts from their lives as argumentative support.

The entire script is required to be output in Modern Chinese. Reasonable artistic license is permitted, but the accuracy of historical facts and philosophical tenets must be ensured. The final output must adhere to standard script formatting requirements.

\subsubsection{Scoring Rubric}

{\small
\begin{longtable}{p{9cm}>{\centering\arraybackslash}p{1.2cm}>{\centering\arraybackslash}p{2cm}>{\centering\arraybackslash}p{2.8cm}}
\caption{Scoring rubric for the HSS example task (Part 2).}
\label{tab:rubric-hss-2} \\
\toprule
\multicolumn{1}{c}{\textbf{Criterion}} & \textbf{Points} & \textbf{Importance} & \textbf{Tag} \\
\midrule
\endfirsthead

\multicolumn{4}{l}{\small\textit{Table~\ref{tab:rubric-hss-2} (continued)}} \\
\toprule
\multicolumn{1}{c}{\textbf{Criterion}} & \textbf{Points} & \textbf{Importance} & \textbf{Tag} \\
\midrule
\endhead

\bottomrule
\endfoot
Content accuracy and depth, restoration of Confucius's philosophical thought: Accurately and three-dimensionally present Confucius's view on life and death, integrating the pragmatism of ``not yet knowing life, how can one know death,'' the value pursuit of ``hearing the Way in the morning, one can die content in the evening,'' and the tension of ``fearing Heaven's mandate while doing one's best,'' with ``benevolence'' and ``ritual'' as the fundamental starting point of all his thoughts and actions. & 10.0 & Essential & Professionalism \\
\midrule
Qualification review (threshold requirement), comprehensive compliance review: This checks whether the work fully meets the basic requirements in format, language, content setting, and factual accuracy. The work must simultaneously satisfy all of the following conditions: 1.\ Format completeness. 2.\ Language standardization (prohibition of semi-classical/modern Chinese mixed with classical Chinese, etc.). 3.\ Premise compliance (Confucius and Socrates meeting in their final years and the ``ignoring temporal/spatial/language'' setting). 4.\ Historical accuracy, no tampering with key historical facts. 5.\ Dialogue balance, presented as bidirectional, back-and-forth dialogue, rather than continuous one-way responses or lecture mode. & 10.0 & Essential & Format Structure \\
\midrule
Content accuracy and depth, restoration of Socrates's philosophical thought: Accurately present his core idea that ``the unexamined life is not worth living,'' treating ``death'' as the ultimate test of ``care of the soul,'' and embodying his unique method of argumentation (maieutics). & 10.0 & Essential & Professionalism \\
\midrule
Content accuracy and depth, use of historical background and events: Ability to use the macro situation of both figures being at ``the turning point from prosperity to decline, seeking the Way but not meeting it'' as the narrative tone, accurately employing historical facts such as ``Western hunt capturing the qilin'' and ``Athenian trial,'' making them internal drivers of the philosophical debate rather than merely external plot devices. All historical details (characters, time) are accurate. & 10.0 & Essential & Accuracy \\
\midrule
Core task completion, effectiveness and depth of intellectual confrontation: Truly achieves intellectual confrontation and collision, with both sides' arguments able to question each other and progress layer by layer, rather than ``talking past each other.'' The debate can go from the surface of views on life and death to the fundamental differences behind views on Heaven's mandate, truth, soul, and social responsibility. & 10.0 & Essential & Logicality \\
\midrule
Core task completion, logical progression of debate: The debate process is logically clear, with questions interlocking, naturally transitioning from one argument to the next, ultimately leading to a meaningful dramatic climax or philosophical reflection. If characters undergo conceptual changes, clearly present the continuous trajectory ``from evidence to psychological inference to behavioral/linguistic representation.'' & 10.0 & Essential & Logicality \\
\midrule
Core task completion, restoration of Socrates's ``maieutics'' style: Accurately reproduces Socrates's ``maieutics'' style of guiding deep thinking through continuous questioning and revealing contradictions in the other's views. & 10.0 & Essential & Professionalism \\
\midrule
Core task completion, restoration of Confucius's ``heuristic'' dialogue style: Accurately reproduces Confucius's heuristic style of being concise, using many metaphors, teaching according to aptitude, aiming to ``draw the bow without shooting,'' emphasizing ``teaching by words and deeds,'' full of authority and moral care. & 10.0 & Essential & Professionalism \\
\midrule
Core task completion, extension of thought and ending treatment: The ending is not just the conclusion of the debate but also the sublimation of thought. Creatively proposes a profound consensus of ``harmony in diversity,'' or uses an open ending to reflect the continuity and complexity of East-West civilizational dialogue, reasonable and with lingering aftertaste. & 2.0 & Optional & Innovation \\
\midrule
Dramatic construction and narrative, plot design and dramatic conflict: Plot design is clever, dramatic conflict is entirely driven by internal intellectual contradictions, and the climax (such as dual death events) greatly strengthens the philosophical theme rather than being superficial. & 10.0 & Essential & Innovation \\
\midrule
Dramatic construction and narrative, character development (main and supporting): Main characters are three-dimensional and full, with thought and emotion interwoven. If there are supporting characters, their words and actions effectively serve main character development and theme expression rather than being tools; characters' actions, expressions, and behaviors are highly consistent with their respective identities and situations. & 10.0 & Essential & Completeness \\
\midrule
Dramatic construction and narrative, functionality of scene setting: Scene setting is not just background but can act as a ``third character'' to drive dialogue, with symbolic elements effectively reflecting characters' inner states or philosophical conflicts. & 10.0 & Essential & Innovation \\
\midrule
Dramatic construction and narrative, use of scenes and symbolic imagery: The use of symbolic imagery (jade pendant/qilin/poisoned wine) is exquisite and natural, effectively creating atmosphere and deeply integrating with the philosophical theme, with clear and recognizable metaphorical associations and multiple interpretive spaces. & 2.0 & Optional & Innovation \\
\midrule
Language and format standards, script format professionalism: Format is complete, layout is professional, scene directions (actions, expressions, etc.) are precise and expressive, effectively guiding secondary creation. & 10.0 & Essential & Format Structure \\
\midrule
Language and format standards, use of modern Chinese and literary quality: Language is fluent, concise, and literary, able to accurately convey characters' identities, personalities, and the weight of thought using modern, accessible language. & 10.0 & Essential & Language Expression \\
\midrule
Language and format standards, avoiding inappropriate ``classical style'': Fully follows modern Chinese requirements, skillfully creating classical charm through word choice and sentence structure rather than simply piling up classical Chinese vocabulary. & 10.0 & Essential & Language Expression \\
\bottomrule
\end{longtable}
}
% Appendix的图片从1开始编号
\counterwithin{figure}{section} % 图片编号依赖于附录章节
\setcounter{figure}{0} % 重置图片计数器

\end{document}